\documentclass[lettersize,journal]{IEEEtran}
\usepackage{caption}
\usepackage{amsmath,amsfonts}
\usepackage{algorithmic}
\usepackage{algorithm}
\usepackage{array}
\usepackage{textcomp}
\usepackage{stfloats}
\usepackage{url}
\usepackage{verbatim}
\usepackage{graphicx}
\usepackage{cite}
\usepackage{enumitem}
\usepackage{subfigure}
\usepackage{xcolor}
\begin{document}

\title{Zero-touch realization of Pervasive Artificial Intelligence-as-a-service in 6G networks}

\author{
    \IEEEauthorblockN{Emna~Baccour\IEEEauthorrefmark{1}, Mhd~Saria~Allahham \IEEEauthorrefmark{2}\IEEEauthorrefmark{3}, Aiman~Erbad\IEEEauthorrefmark{1}, Amr~Mohamed\IEEEauthorrefmark{2}, Ahmed Refaey Hussein\IEEEauthorrefmark{4}, and Mounir~Hamdi\IEEEauthorrefmark{1}.
    }
 \thanks{\IEEEauthorrefmark{1} E. Baccour, A. Erbad, and M. Hamdi are with the College of Science and Engineering, Hamad Bin Khalifa University, Qatar Foundation.}
\thanks{\IEEEauthorrefmark{3} M.S. Allahham is now with the School of Computing, Queen’s University, ON, Canada. This work was done while he was a research fellow at Qatar University.}
\thanks{\IEEEauthorrefmark{2} A. Mohamed is with the
Department of Computer Science and Engineering, College of Engineering, Qatar University, Qatar.}
\thanks{\IEEEauthorrefmark{4} A. R. Hussein is with the School of Engineering, University of Guelph, ON, Canada.}
\vspace{-0.6cm}
}


\maketitle

\begin{abstract}
The vision of the upcoming 6G technologies, characterized by ultra-dense network, low latency, and fast data rate is to support Pervasive AI (PAI) using zero-touch solutions enabling self-X (e.g., self-configuration, self-monitoring, and self-healing) services. However, the research on 6G is still in its infancy, and only the first steps have been taken to conceptualize its design, investigate its implementation, and plan for use cases. Toward this end, academia and industry communities have gradually shifted from theoretical studies of AI distribution to real-world deployment and standardization. Still, designing an end-to-end framework that systematizes the AI distribution by allowing easier access to the service using a third-party application assisted by a zero-touch service provisioning has not been well explored. In this context, we introduce a novel platform architecture to deploy a zero-touch PAI-as-a-Service (PAIaaS) in 6G networks supported by a blockchain-based smart system. This platform aims to standardize the pervasive AI at all levels of the architecture and unify the interfaces in order to facilitate the service deployment across application and infrastructure domains, relieve the users worries about cost, security, and resource allocation, and at the same time, respect the 6G stringent performance requirements. As a proof of concept, we present a Federated Learning-as-a-service use case where we evaluate the ability of our proposed system to self-optimize and self-adapt to the dynamics of 6G networks in addition to minimizing the users' perceived costs.
\end{abstract}
\section{Introduction}
Responding to the ever-increasing advancements of our smart digital world, 6G has been introduced to modernize a plethora of intelligent services and innovative applications by promptly deploying advanced AI anytime and anywhere~\cite{8808168} across the worldwide network. The support of the AI-empowered services is one of the key missing elements in previous generations and will be the most significant driving force in the 6G leap. 
The intelligent interactions between devices across the network called for Pervasive Artificial Intelligence (PAI), defined as \textit{“The intelligent and efficient distribution of AI tasks and models over/amongst any type of devices with heterogeneous capabilities in order to execute sophisticated global missions”}~\cite{DBLP:journals/corr/abs-2105-01798}. The PAI that includes Federated Learning (FL) and distributed inference can, on the one hand, take advantage of the ultra-quality of the 6G network, the massively interconnected devices, and the explosive amount of generated data. On the other hand, it can empower the promises of the 6G to provide an intelligent communication environment characterized by large-scale automation and hence achieve a smooth evolution from people connection to  \textit{"connected intelligence"} \cite{8808168}.


The confluence of 6G and PAI relies basically on the pervasive, intelligent, and seamless cooperation between a massive number of heterogeneous devices executing complex tasks requested by diverse service consumers. This clearly calls, first, to conceive a structured approach that efficiently manages heterogeneous end-users demands by designing the future 6G network as a chain of PAI services. Second, an innovative and trusted solution to adapt to the dynamics of these services and optimize the scarce resources has become an urgent necessity. This solution should be capable of self-monitoring, self-optimizing, and self-healing with high reliability and minimum human intervention. In this context, the \textit{Zero-touch Network and Service Management (ZSM)} concept~\cite{GALLEGOMADRID2022105} has been recently introduced to automatically orchestrate network resources using emerging techniques of machine learning (e.g., reinforcement learning), game theory, and optimization, among others, aiming to pursue the 6G key performance indicators. 

Motivated by the emergence of automated services, we propose a Zero-touch realization of Pervasive Artificial Intelligence-as-a-Service (PAIaaS) in 6G networks. Specifically, we design a new horizontal and vertical end-to-end architecture framework achieving a closed-loop trusted automation of the network and service management to ensure the economic sustainability and high agility of the diverse set of PAI services. This framework aims to standardize the pervasive AI at all levels of the architecture and unify the interfaces to facilitate service deployment across different application and infrastructure domains. This systematizes the technology and makes the distribution, security, and resource allocation process seamless to the end-users, infrastructure providers, and pervasive devices, to rather focus on other domain-specific details. Some related papers, including \cite{10.1145/3426745.3431337} and \cite{s20205796}, have recently proposed distributed AI services. However, they did not present an end-to-end architecture enclosing the whole process, nor they envisaged a zero-touch trusted management of the service provisioning. In this work, we present a high-level overview of the proposed zero-touch PAI-as-a-service complemented by a detailed description of different architectural elements and their automated interactions, starting from consumer requests to service delivering. Our main contributions are described as follows:
\begin{itemize}
\item We propose a novel solution to deploy zero-touch PAI-as-a-Service in 6G networks.
\item We present a reference pillar architecture for the PAIaaS and detail the interaction between different stacked elements, i.e., service consumer, Application Service Provider (ASP), infrastructure, and pervasive system.
\item Because of the dynamics of the network and the need for decentralization and reliability of the PAI services, we propose to use blockchain to model the  Smart Contracts (SCs) between different elements of our framework and control their reputations.
\item For an adaptive blockchain solution, we propose to assist the SCs with a Deep Reinforcement Learning (DRL) approach that guarantees self-tuning resource assignment, self-monitoring of components' reputations, and self-adaptation to any fluctuation in the environment. 
\item As a proof of concept, we evaluate the PAIaaS potential by simulating a federated learning use case.

\end{itemize}
\section{Zero-touch PAIaaS architecture}
PAIaaS aims to provide to end-users with no technical knowledge an easy and transparent way to use pervasive AI techniques without any effort or cost of developing, tuning, and testing the algorithms. Another objective of the PAIaaS is to meet the real-time and Quality of Service (QoS) requirements, owing to the agreements with the 6G infrastructure providers to make their resources available on demand. In this way, the services can be promptly deployed once customized to the needs of the end-user. 
On the other hand, the PAIaaS can help the infrastructure providers enlarge their businesses and reach a wider clientele by systematizing the PAI services and delegating their management to another automated component of the system.

\begin{figure}[h]
\centering
	\includegraphics[scale=0.26]{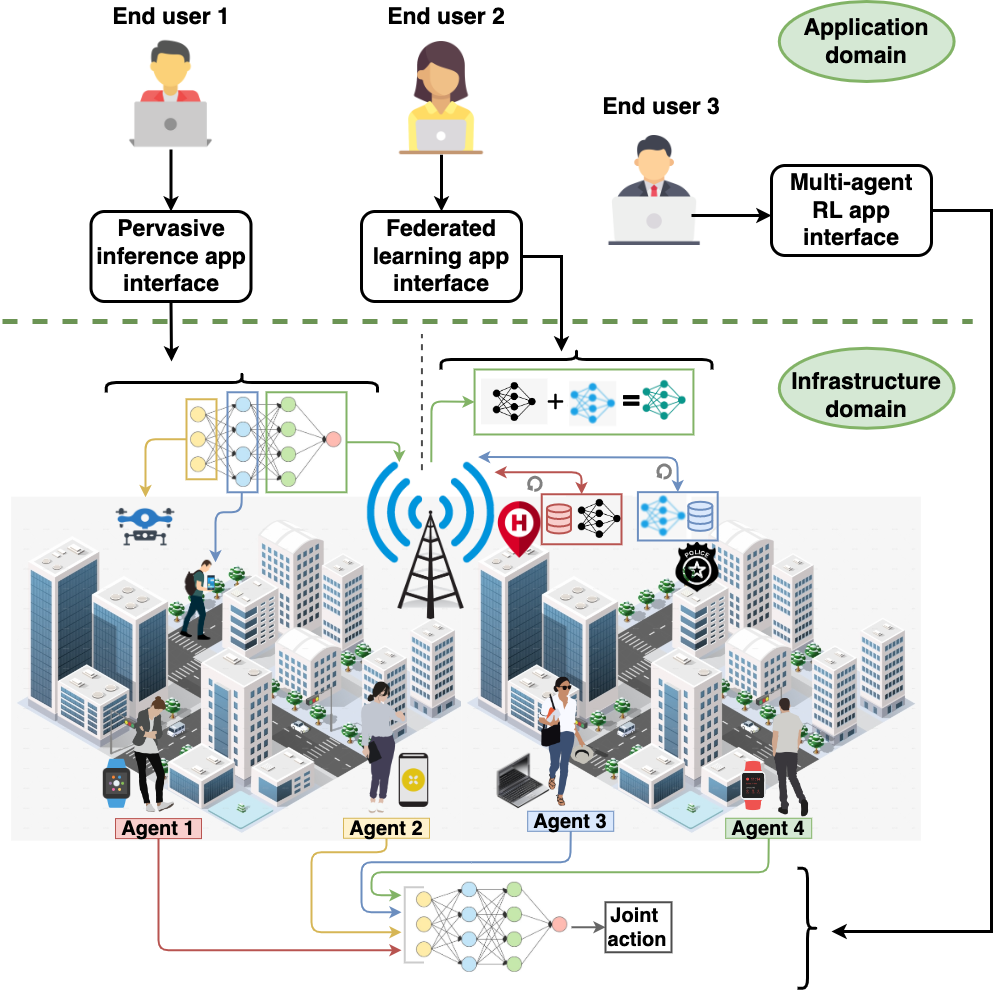}
	\caption{PAIaaS high-level domains.}
	\label{domains}
\end{figure}

From a high-level point of view, the PAIaaS framework can be administratively and functionally split into two domains, as illustrated in Fig. \ref{domains}: (1) The application domain that hosts the front-end part allowing the end-user to choose the service and customize it, and the controller part that connects to the infrastructure level in order to implement the PAI algorithms. (2) The infrastructure domain, which includes the facilities to deploy the applications, the infrastructure services (e.g., data, communication, and storage.), and the ubiquitous devices that participate in the computation of the AI sub-tasks.

\begin{figure*}[h]
\centering
	\includegraphics[scale=0.45]{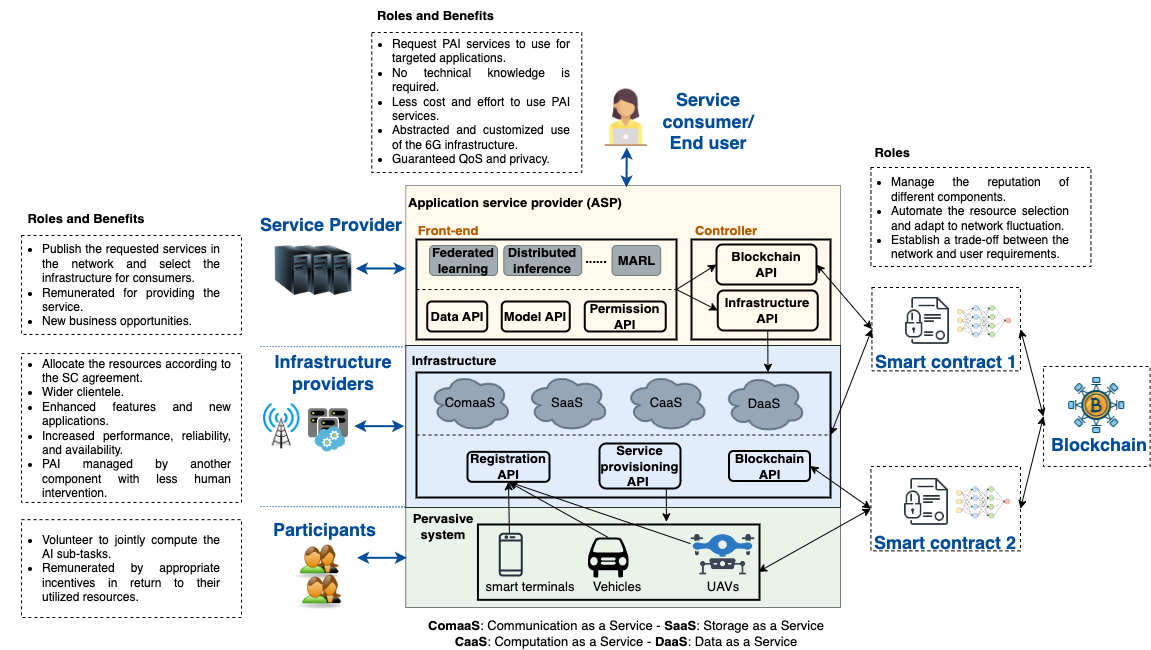}
	\caption{PAIaaS framework stack.}
	\label{framework}
\end{figure*}

Our goal is to design the conceptual framework of the pervasive AI-as-a-service and correspondingly propose a software that is able to systematize the relation and trust between different components and automate their tasks, aiming at achieving the zero-touch PAI vision as a whole. Therefore, we propose a multi-tiered approach illustrated in Fig. \ref{framework}, where at a high-level, our pillar framework comprises the service consumers that benefit from the PAI services offered by the Application Service Provider (ASP). The ASP employs the infrastructure facilities to eventually enable the requested services. At a lower level, the pervasive participants share their information with the infrastructure provider in order to enroll them as computation/storage resources. Finally, the software that automates the communication and trust between all these components represents a smart contract agent modeled as a blockchain reputation system and controlled by reinforcement learning. The roles of different components and their benefits from the PAIaaS framework are summarized in Fig. \ref{framework}. A detailed description is presented in what follows.
\subsection{End-user/service consumer}
A service consumer is a user that potentially owns a project or a product, such as system administrator (e.g., surveillance system.) or software expert (e.g., music application) that does not typically possess any technical skills to understand the offered services or simply does not have the means to deploy the PAI system. 
End-users may request different types of services, including the ones illustrated in Fig. \ref{domains}:
\begin{itemize}[leftmargin=*]
\item \textbf{Federated Learning-as-a-Service (FLaaS):}
To preserve privacy and reduce the complexity of training a huge dataset, FL end-users can engage a server through the PAIaaS framework to aggregate the global model after receiving different trained models at locally allocated machines. The FLaaS users can be, for example, independent hospitals' administrators that own medical data and wish to train a global model on a higher diversity of observations, while preserving the privacy of their patients. 
\item \textbf{Distributed inference-as-a-Service (DIaaS):} 
To cope with limited edge resources and simultaneously avoid latency overheads caused by cloud transmissions, the inference tasks can be distributed among ubiquitous devices located in the proximity of the data source.
A potential DIaaS consumer is a bank or an airport  surveillance system that captures a tremendous amount of high-resolution images for monitoring purposes. This system may wish to exploit the collaborative learning due to shortage of local resources.
\item \textbf{Multi-agent Reinforcement Learning-as-a-Service (MARLaaS):} Some AI techniques are inherently distributed, such as Multi-agent Reinforcement Learning, where agents cooperate to build and improve a policy in real-time, enabling them to take on-the-fly decisions/actions based on the environment status. Potential users for MARLaaS can be a set of servers (e.g., Spotify servers or online shopping websites) that run a recommender system for their prospective clients. The goal of each one is to recommend the most popular content across all servers (e.g., songs, videos, and books.) without violating others' privacy. 
\end{itemize}
\subsection{Service provider (SP)}
The Service Provider Application (APS) should enable the end-users to benefit from the services without any complexity or understanding of the underlying technology. Currently, the existing PAI libraries such as TFF~\cite{tensorflow_2020} designed for FL do not present a service model or a GUI that can be easily used by service consumers. Moreover, commercial Machine Learning-as-a-Service (MLaaS) applications (e.g., AWS \cite{amazon_2021}), that provide graphical user interfaces, do not support PAI services. Therefore, the PAIaaS ASP will provide RESTful APIs to support a model building exposing useful abstractions for the end-users. The model building design comprises two components:
\subsubsection{Front-End} presents the main interface (or the GUI) for the service consumers, enabling them to bootstrap, choose the service, configure, and exit after the termination. The GUI uses a set of APIs under the hood, which are: 
\begin{itemize}[leftmargin=*]
    \item \textbf{Data APIs:} it allows the users to input all required details about the data, e.g., type, size, or shape of the environment's states for MARL. These data can be inserted as a JSON format, where each input is defined by a name and a type.
    \item \textbf{Model APIs:} it allows the users to specify their ML models (e.g., VGG for distributed inference.), parameters (e.g., exploration rate for MARL), and algorithms (e.g., FedAvg for FL) either by creating new ones and customizing them or by choosing built-in PAI models.
    \item \textbf{Permission APIs:} it allows the end-users to give access to other contributors to use the same parameters or join the collaborative system. In the Federated Learning scenario, one of the participants can create the initial model and invite the other data owners to join the learning process.  
\end{itemize}
After receiving the customer request, the front-end component calls the functions on the controller to execute it.
\subsubsection{Controller} 
it takes the inputs inserted by the user in the front-end component to configure the services, e.g., initialize the trained model, initialize the algorithms, and set the granted permissions.  Once the service starts, the controller is responsible for selecting the appropriate infrastructure providers (i.e., data, communication, computation, etc.) based on the requirements of the service, the QoS/reputation presented by different providers, their cost, and their availabilities. The smart contract agent (detailed in section \ref{sc}) will be in charge of this selection task along with the service monitoring to adapt to any change or incident. Hence, two APIs are required:  
\begin{itemize}[leftmargin=*]
    \item \textbf{Blockchain API (e.g., Web3.js):} it allows the controller to interact with a remote node (e.g., Ethereum.) using HTTP, IPC, or WebSocket in order to create a blockchain smart contract evaluating the reputation and performance of the infrastructure providers before selection. 
    \item \textbf{Infrastructure API:} the customized service needs to be embedded in an API/ software library that (1) authenticates and provides permissions for the selected Infrastructure Providers (IPs) to access the user’s data, (2) provides the necessary PAI functions to execute the needed tasks, and (3) gives the permission to exchange/share data with the end-user.
\end{itemize}
Finally, to assure the privacy of users’ data, it is critical to leverage privacy-preserving mechanisms, such as Differential Privacy (DP) \cite{9253545} while sharing any information to external components, e.g., IPs or IoT participants. Adding noise to the AI data may affect the performance of the models; hence, we can build our work on existing DP solutions \cite{9253545}.
\subsection{Infrastructure Providers (IPs)}
Although PAI has recently gained enormous attention, many challenges are still facing its realistic deployment in current 5G networks. Specifically, 5G networks are recognized for their capacity to revolutionize a diversity of IoT services and applications. However, their era coincided with the revolution of AI that has driven machine intelligence towards full autonomy to operate smart city applications. In practice, the network latency still poses a problem for AI services, particularly PAI, due to the communication between participants that adds an overhead to the service perceived time. Moreover, the system should support a high density of connected devices, high heterogeneity, context-awareness, and seamless access to computing whenever needed. Since the current networks are insufficient to fully satisfy the stringent requirements of newly developed forthcoming services, the efforts from academia and industry started to conceptualize 6G networks with special attention to emerging distributed AI workloads. Achieving a better performance than 5G is a critical step, as it is not sufficient to expand spectrum capacity and implement new technologies, but also the ambitious energy reduction goals should be met \cite{peeters_stern_ramirez_izumo_russo_2021,8808168} (see Table \ref{6G}).

\begin{table}[!h]
\caption{ Novel 6G characteristics empowering PAI.}
\resizebox{\columnwidth}{!}{%
\begin{tabular}{|ll|}
\hline
\multicolumn{1}{|c|}{\textbf{5G}} & \multicolumn{1}{c|}{\textbf{6G}} \\\hline
\multicolumn{2}{|c|}{\textbf{Technologies}} \\ \hline
\multicolumn{1}{|l|}{\begin{tabular}[c]{@{}l@{}}-\textbf{Ultra-Reliable and Low-Latency}\\\textbf{Communications (URLLC).} \\ \\ -\textbf{Enhanced Mobile BroadBand}\\\textbf{(eMBB).} \\ \\  -\textbf{Massive Machine–type} \\\textbf{Communications (mMTC).} \\ \\ \\  -\textbf{Antenna architecture:} MIMO\\ \\ -\textbf{Satellite integration:} no\\ \\ -\textbf{Wireless power transfer:} no\\ -\textbf{AI integration:} partial\\ \\ -\textbf{Smart city components:} separate\end{tabular}} &\begin{tabular}[c]{@{}l@{}}-\textbf{Event Defined uRLLC (EDuRLLC):} \\ supports extreme events such as fluctuating\\ device densities.\\ -\textbf{Contextually Agile eMBB Communications} \\\textbf{(CAeC):} eMBB are adaptive to the network\\   context, e.g.,congestion, and social context.\\ -\textbf{Computation Oriented Communication}\\ \textbf{(COC)}: flexible selection of communication\\ resources to achieve computational accuracy \\ for learning approaches.\\ -\textbf{Antenna architecture:} Reconfigurable \\ Intelligent Surfaces (RISs).\\ -\textbf{Satellite integration:} Always-on network \\coverage.\\ -\textbf{Wireless power transfer:} Yes\\ -\textbf{Connected intelligence:} fully deploying AI\\ capabilities\\ -\textbf{Smart city components:} integrated\end{tabular} \\ \hline
\multicolumn{2}{|c|}{\textbf{Characteristics}} \\ \hline
\multicolumn{1}{|l|}{\begin{tabular}[c]{@{}l@{}}-\textbf{Connection density/$km^2$:} $10^6$\\ -\textbf{Data rate:} 20 Gb/s\\ -\textbf{Latency:} 1 ms\\ -\textbf{Frequency:} 3-30 GHz\\ -\textbf{Energy consumption:} 1000 $\times$ \\ better than 4G\\ -\textbf{Network service area:} 70\%\\ -\textbf{Mobility:} 500 km/h\end{tabular}} & \begin{tabular}[c]{@{}l@{}}-\textbf{Connection density/$km^2$:} $10^7$\\ -\textbf{Data rate:} 1 Tb/s\\ -\textbf{Latency:} 10–100 $\mu$s\\ -\textbf{Frequency:} 95 GHz-3 THz\\ -\textbf{Energy consumption:} supports battery\\  free devices, 10 $\times$ better than 5G \\ -\textbf{Network service area:} more than 90\%\\ -\textbf{Mobility:} 1000 km/h\end{tabular} \\ \hline
\end{tabular}}
\label{6G}
\end{table}
On the one hand, the 6G is proposed in our framework design to enable  high-performance PAI services. On the other hand, PAaaS can play a crucial role in achieving the vision of the upcoming 6G (e.g., AI-empowered services.). On these bases, agreements between the service provider and 6G infrastructure providers should be established to offer on-demand resources (e.g., communication and data services.).

Similarly to the application level, multiple APIs are designed to unify the interaction with the pervasive system:
\begin{itemize}[leftmargin=*]
    \item \textbf{Registration API:} it allows the pervasive devices to volunteer for resource provisioning by registering and sharing their information, including their ids, available resources, and dynamic locations. During registration, since the participants are highly heterogeneous and may use different computation technologies that need to be unified, the infrastructure provider should instruct the required PAI standards and provide the mechanisms and tools supported by each type of device. These standards and tools are selected and managed by the corresponding infrastructure provider.
    \item \textbf{Blockchain API:} the blockchain SC agent negotiates the required resources for the service against the constraints of the end-user and the IPs (e.g., cost and load balancing constraints). Upon agreement, the infrastructure providers authenticate the selected devices for the service.
    \item \textbf{Service provisioning API:} serves to give IoT devices access to user’s data, and PAI functions and algorithms. Moreover, PAIaaS should provide mechanisms to grant the IoT devices permission to exchange data with end-users as well as the tools to apply privacy mechanisms on the shared data.
\end{itemize}
\subsection{Pervasive system}
The pervasive system comprises the nodes sharing resources under their control after registering to the infrastructure provider, retained through suitable incentives. Their contribution is, therefore, managed at a higher-level through the blockchain-based RL system. The pervasive devices can span from resource-constrained devices to high-performance servers and can involve cloud data centers, mobile edge computing servers, mobile devices, wearable computers, tablets, and even a TV. These ubiquitous devices are constantly connected and available for any task. 

\subsection{Blockchain-based smart contract agent}\label{sc}
The pervasive devices are potentially not commercialized for resource allocation tasks. This means they are not able to optimize and manage their resources by design, which restrains the system from being autonomous and self-configured. Therefore, a reliable resource management framework should be designed. Since the pervasive system is highly dynamic and decentralized, blockchain is a promising candidate \cite{9706476} owing to its distributed and trusted architecture. 
In fact, the blockchain does not store the shared information in a centralized location. Instead, the data is copied and distributed among decentralized devices spread all over the network. When a new block is generated, every node updates its blockchain. In this way, it is difficult to tamper with the data, as a hacker cannot compromise all the decentralized copies, and any malicious attack is easily detected by checking all nodes. Furthermore, the blockchain system is complemented by an efficient payment process that can approve the transactions and immediately trigger the payment using cryptocurrency without engaging a third-party payment system.
Another advantage of blockchain is the possibility to deploy any software application or programmed logic with multiple appealing security guarantees, namely Smart Contract (SC) \cite{Wood2014ETHEREUMAS}. The SC can be defined as a digital agreement between the system components, and it is characterized by its immutable and deterministic code that is automatically executed when all the required conditions are met without the intervention or the control of another party, which matches the requirements of our Zero-touch system. These SC's proprieties are also essential to gain the trust of different components that may feel reluctant to apply non-autonomous negotiation frameworks. Recently, multiple applications started to use the blockchain SC for resource management problems, including the work in ~\cite{9573346}. 

In our context, SCs are applied for service provisioning problems, where multiple factors should be considered, e.g., the service cost that the end-users pay and the resources load that the infrastructure and the pervasive devices endure in addition to their reputations. 
In general, such task allocation contracts are solved using joint optimization. However, solving an optimization using classical approaches in 6G networks is computationally expensive, if not impossible, due to the scale, density, and heterogeneity. Moreover, these approaches do not consider the dynamics of the system over time.
It means that finding a current optimal solution might not be valid for future system status, which calls for adaptive and  self-optimized blockchain SCs.

In our PAIaaS zero-touch system, we propose to implement an intelligent SC that not only guarantees a secure and trusted service provisioning but also an adaptive and self-tuning resource assignment to establish a trade-off between users' requirements and providers' constraints. This intelligent SC is based on learning-assisted decision-making, namely Deep Reinforcement Learning (DRL). The DRL sets up a feedback loop between the decision maker and the physical system in order to refine the actions based on the environment dynamics until reaching eventual optimality.

Therefore, we envisage two intelligent SCs, as illustrated in Fig. \ref{framework}. The first SC, created by the SP, selects the infrastructure providers that are the most appropriate for the requested service based on the task requirements, incurred costs, load, and reputation. The second SC is deployed in the infrastructure domain, where the decision will be the selection of the available and cost-effective IoT pervasive devices belonging to the designated IPs. We note that the service provider collects the PAI tasks in the form of transactions and submits them for computation. Therefore, we define $\mathcal{TX}=\{1,2,…,TX\}$ as the set of transactions assigned to the infrastructure providers (e.g., inference requests.), and then split and distributed to the pervasive system in the second SC resource management problem (e.g., segments of a CNN inference.). We also denote $\mathcal{R}=\{1,2,…,R\}$ as the set of resource providers (IPs in the first SC or the set of IoT devices in the second one). It is worth highlighting that the blockchain is, by design, agnostic to the type of participants. In our context, the structures of both smart contracts are the same as they are both deployed in blockchain and assisted by DRL. Still, they have different functions since they control the relation between different components in different environments. 

In our formulation, we model the environment composed of resource providers and transactions requested by end-users as a Markov Decision Process (MDP). The MDP is generally defined as a tuple $(S,A,P,R,\gamma)$, where $S$ denotes the set of states, $A$ is the action space, $P$ is the transition probability, $R$ is the reward, and $\gamma$ is the discount factor. In MDP RL, the action $A$ affects not only the immediate reward but also the next environment’s states and potentially all subsequent rewards. This allows an efficient assignment of the available resource providers while considering future demand. As such, the RL policy $\pi$ is assisted by the blockchain model to be learned. Specifically, the chained data contains indications about the performance of previous service provisioning (i.e., reward) and participants review in the form of experience (i.e., $<$$s_t,a_t,r_t,s_{t+1}$$>$). Given the previous experiences, the service provisioning policy can be enhanced in real-time at each new block until maximizing the rewards. We note that the reward function reflects the objective of maximizing the reputation score and optimizing the requirements of the end-users and IPs (e.g., cost and load). The temporal structure of blockchain-based systems (i.e., blocks periodically generated and appended) makes them excellent candidates to leverage online-learning algorithms, since SCs are envisioned as autonomous entities that learn from the chained data. We also highlight that the size of data stored in each block (e.g., states of devices and transactions.) is not large and does not burden the system with unreasonable storage and communication requirements. More precisely, independently of the system's size, only a small and sufficient batch of connected devices is assigned for each PAIaaS request, which limits the size of states stored in each block. 

\begin{figure}[h]
\centering
	\includegraphics[scale=0.38]{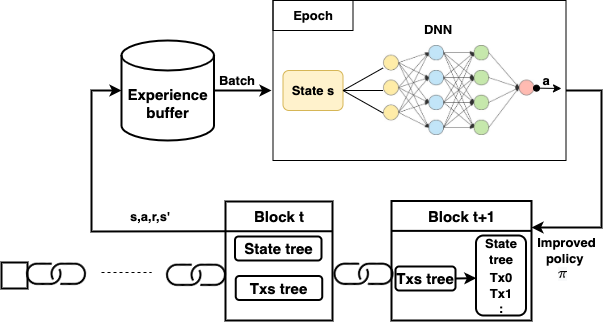}
	\caption{Blockchain-based system.}
	\label{bolckchain}
\end{figure}
\begin{figure*}[h]
\centering
	\includegraphics[scale=0.44]{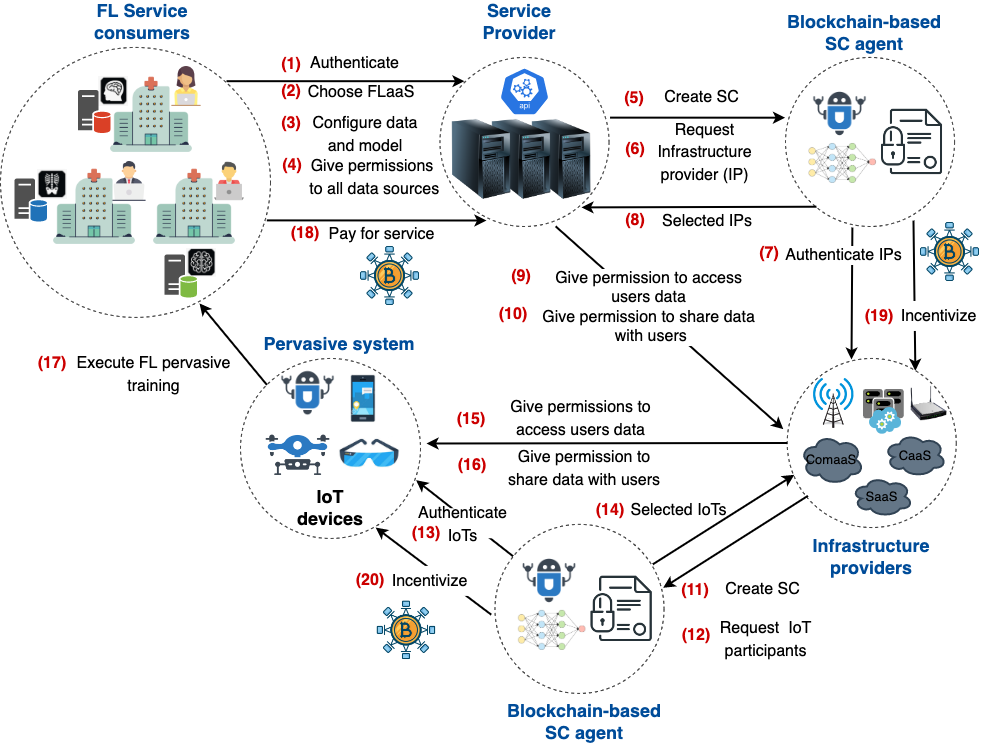}
	\caption{Federated Learning use case: sequence diagram.}
	\label{seq}
\end{figure*}
The blockchain-based system is illustrated in Fig. \ref{bolckchain}, and different components of the RL system are defined as follows:
\begin{itemize}[leftmargin=*]
\item \textbf{States space:} At each time step $t$, a service transaction request $t_x \in \mathcal{TX}$ should be assigned to the resource provider $r \in \mathcal{R}$.
The SC agent takes the decision based on the set $s_t=\{g_t, d_t, l_t, c_t, f_t\}$, where $g_t$ is the type of the task (e.g., computation, storage, and data.), $d_t$ is the resource demand of the transaction $t_x$, $l_t$ represents the vector of loads of different resource providers, $c_t$ denotes the vector of costs for serving the transaction (e.g., energy, computation, incentive fees.). We note that the incentive-based mechanism is out of our scope and we can build our system on existing mechanisms \cite{BACCOUR2020102801}. Finally, $f_t$ presents the vector of reputation scores. 
\item \textbf{Action space:} Based on the set of states at the step $t$, the SC agent takes an action $a_t$ presenting a vector of binaries, where $a_t^{(i)}=1$ means that the resource $i$ is selected for the transaction. Note that it is possible to deliver the service by multiple collaborative service providers.
\item \textbf{Reward function:} The reward should match the high-level objective to be maximized, which depends on the end-user requirements (e.g., cost) and IPs constraints (e.g., load-balancing). This reward can be formulated as follows:
\begin{equation}\label{reward2}
r_t=C^{(i)} \times [\underbrace{f_t^{(i)} \times (1-c_t^{(i)})}_{T_1}+\underbrace{\frac{1}{R}\sum_{j=0}^R(1-l_t^{(j)})}_{T_2}],
\end{equation}
where $C^{(i)}$ is equal to 1 if the selected device $i$ has the required resources $d_t$ for the type of service $g_t$, and it is not fully loaded; 0 otherwise. $T_1$ contributes to minimizing the service cost by maximizing $(1-c_t^{(i)})$. It also ensures the selection of the device with the best reputation $f_t^{(i)}$. The reputation can indicate the average time a resource provider takes to finish a transaction. This variable depends on the capacity of the provider. In other words, the devices with higher capacities will receive better feedback. Finally, $T_2$ contributes to balancing the load of the resource providers by minimizing their average load.

\end{itemize}
\section{Proof-of-concept: Federated learning use case}
To prove the performance of our zero-touch PAIaaS platform, we introduce a federated learning use case, where the service consumer uses the service provider application to deploy distributed training on the 6G infrastructure relying on pervasive devices. Specifically, storage/computational devices are needed to train local data in addition to the centralized server for aggregation. Communication services are also allocated to share the local models. Fig. \ref{seq} shows different steps of the FLaaS use case, using different APIs and tools described in previous sections. In this simulation, we use the Deep Q-Network (DQN) for RL implementation, and we suppose that the pervasive system is composed of 10 devices with CPU capacities in the range of $[0.2 - 0.8]$ GHz and service cost in the range of $[10^{-8} - 10^{-9}]$ U/cycle. Furthermore, the demands for the FL transactions are in the range of $[0.1-2] \times 10^9$ cycles.

\begin{figure}[!h]
	\centering
	\vspace{-5mm}
		\subfigure[\label{reward}]{\includegraphics[scale=0.278]{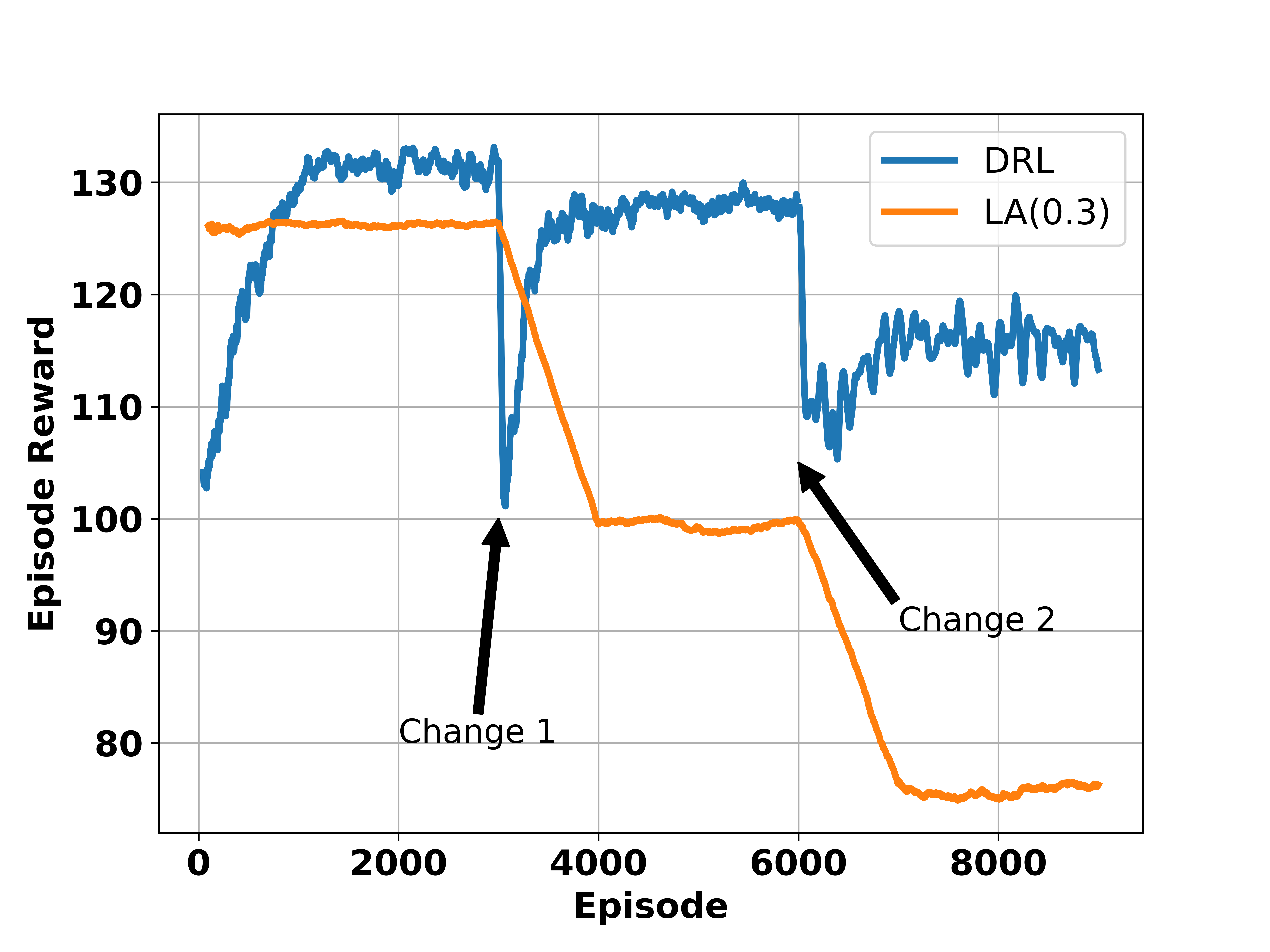}}\\
		\vspace{-3mm}
        \subfigure[\label{cost}]{\includegraphics[scale=0.26]{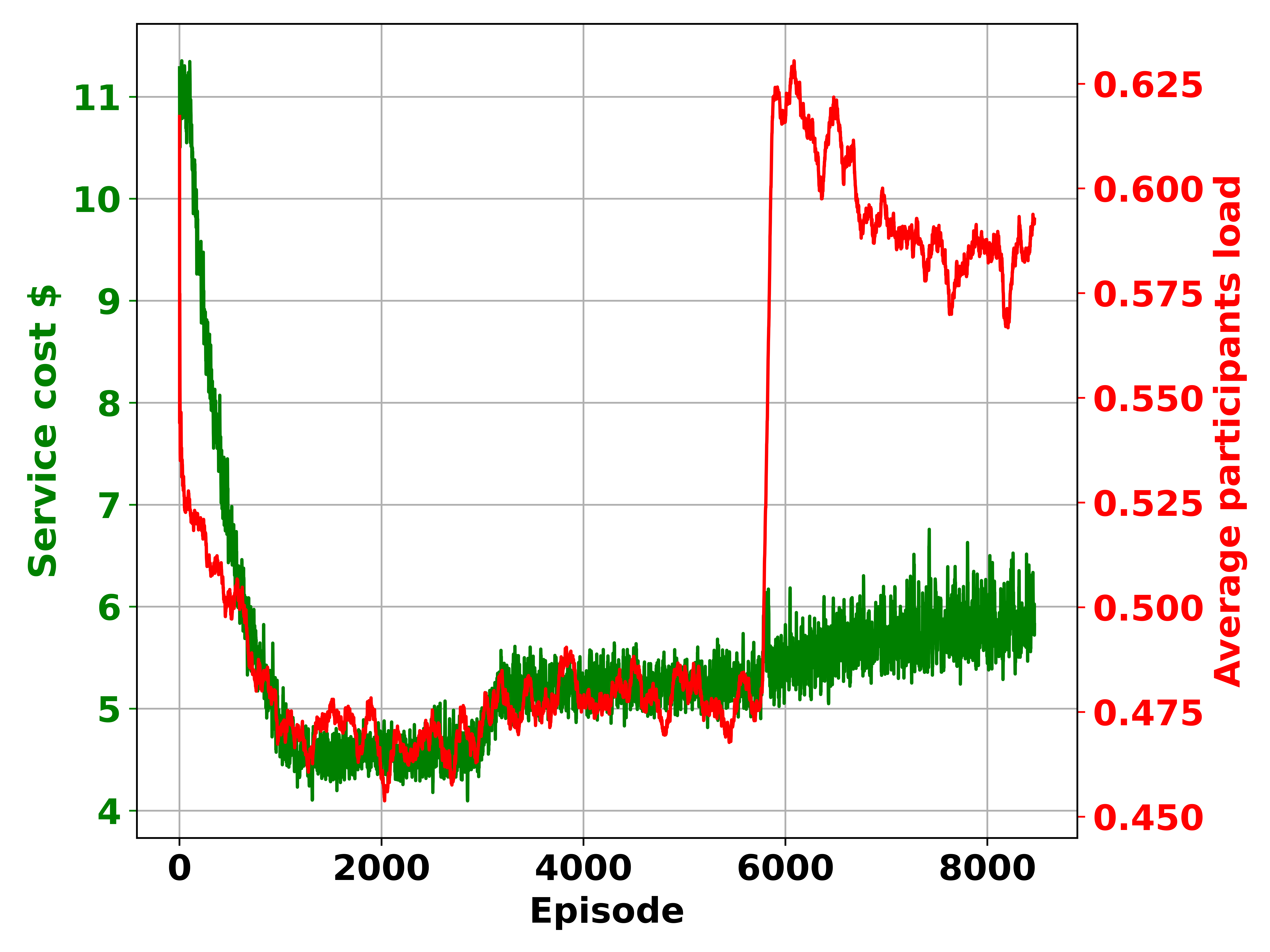}}
        \vspace{-3mm}
	\caption{(a) SC Agent performance after introducing  changes, (b) cost and load incurred by assigned transactions according to the learned policy.}
	\label{RL}
\end{figure}
In this simulation, we first evaluate the performance of our zero-touch platform in terms of self-optimization, self-monitoring, and self-healing. We note that we implemented the smart contract between the IP and the pervasive resources, and we expect that the  smart contract between the SP and the IPs performs identically. As such, we illustrate in Fig. \ref{RL}(a) the obtained rewards while training the SC agent. The first episodes are fully exploratory in order to discover the actions leading to higher rewards. After that, the exploration decays until the actions taken by the SC agent become almost fully exploitatory and stability is reached, which confirms the self-optimization of the system. After the convergence of the system, 3 devices leave the collaborative network as a first change. In this case, the agent experiences a never seen situation, which decreases the performance of the taken actions. However, the SC agent promptly regains its convergence and learns again the new policy relying only on the remaining devices. This rapid re-convergence proves the self-monitoring and self-healing of our proposed platform. The same behavior is also proved when reducing the available resources per device, as a second change. Under the same configuration, the performance of the Load-Aware (LA $\omega$) algorithm is shown. The LA $\omega$ is a rule-based algorithm that assigns the transaction to the service provider with the lowest cost whose load is less than $\omega$. We can see that the Load-Aware performance decreases when changing the environment without adapting to the new changes. Hence, it can be concluded that online and experience-driven learning is of utmost importance in dynamic environments.

Fig. \ref{RL}(b) shows the serving cost (a unit-less parameter that may refer to the consumed resources or incentive fees.) and the average load among devices during the training and after different changes in the system. It can be seen that the SC agent learns the optimal policy slowly until reaching the step where the cost and average load are minimal within all participants. We can also notice that the cost increases when decreasing the number of devices or reducing the available resources due to the overload caused by those changes. Still, the SC agent keeps self-optimizing until achieving a load-balancing welfare. 

\section{Conclusion}
In this paper, we introduced PAIaaS, the first to the best of our knowledge, pervasive AI as a service platform that enables a third-party application to deploy collaborative and zero-touch AI techniques in 6G networks. Particularly, the horizontal and vertical end-to-end architecture framework designed for closed-loop automation of PAI management is presented, where four levels are envisaged: consumers, SPs, IPs, and pervasive system. Those involved actors, their roles, and their interactions using the blockchain-based SC agents, as well as their potential benefits, are identified and discussed. We also presented the proof of concept on a FLaaS use case. Our long-term goal is to create and deploy a prototype of the platform in 6G networks for large-scale evaluation.
\section*{Acknowledgment}
This publication was made possible by NPRP13S-0205-200265 from the Qatar National Research Fund.

\bibliographystyle{IEEEtran}
\bibliography{References.bib}

\newpage
\end{document}